\documentclass{ecai} 

\usepackage{latexsym}
\usepackage{amssymb}
\usepackage{amsmath}
\usepackage{amsthm}
\usepackage{booktabs}
\usepackage{graphicx}
\usepackage{color}
\usepackage{longtable}
\usepackage{booktabs}

\usepackage{cleveref}
\usepackage{paralist}
\usepackage{graphicx}

\usepackage{subcaption} 
\usepackage{multirow}
\usepackage[table,xcdraw]{xcolor}
\usepackage{colortbl}

\newcommand{\BibTeX}{B\kern-.05em{\sc i\kern-.025em b}\kern-.08em\TeX}

\begin{document}


\begin{frontmatter}




\title{Beyond Listenership: AI-Predicted Interventions Drive  \\
Improvements in Maternal Health Behaviours}


\author[A]{\fnms{Arpan}~\snm{Dasgupta}
\thanks{Corresponding Author. Email: arpandg@google.com}\footnote{Equal contribution.}}
\author[B]{\fnms{Sarvesh}~\snm{Gharat}
\footnotemark}
\author[C]{\fnms{Neha}~\snm{Madhiwalla}
}
\author[C]{\fnms{Aparna}~\snm{Hegde}
}
\author[A, D]{\fnms{Milind}~\snm{Tambe}
}
\author[A]{\fnms{Aparna}~\snm{Taneja}
}

\address[A]{Google DeepMind}
\address[B]{IIT Bombay}
\address[C]{ARMMAN}
\address[D]{Harvard University}


\begin{abstract}
Automated voice calls with health information are a proven method for disseminating maternal and child health information among beneficiaries and are deployed in several programs around the world. However, these programs often suffer from beneficiary dropoffs and poor engagement. In previous work, through real-world trials, we showed that an AI model, specifically a restless bandit model, could identify beneficiaries who would benefit most from live service call interventions, preventing dropoffs and boosting engagement. However, one key question has remained open so far: does such improved listenership via AI-targeted interventions translate into beneficiaries' improved knowledge and health behaviors? We present a first study that shows not only listenership improvements due to AI interventions, but also simultaneously links these improvements to health behavior changes. Specifically, we demonstrate that AI-scheduled interventions, which enhance listenership, lead to statistically significant improvements in beneficiaries' health behaviors such as taking iron or calcium supplements in the postnatal period, as well as understanding of critical health topics during pregnancy and infancy. This underscores the potential of AI to drive meaningful improvements in maternal and child health.
\end{abstract}

\end{frontmatter}



\section{Introduction}

Maternal mortality remains a major global health concern, particularly in developing countries, and its reduction is a critical target highlighted by the World Health Organization (WHO) under the Sustainable Development Goals (SDGs) \cite{WHOTargetMaternal}.
Timely access to reliable health information plays a crucial role in improving maternal and infant health outcomes, particularly in underserved communities where traditional healthcare services, and access to resources may be limited. Mobile health (mHealth) programs ~\cite{10.1371/journal.pone.0262842, healthcare10010004} have emerged as an effective way to bridge this gap in several public health domains, leveraging the widespread use of basic mobile phones 
to deliver essential health education at scale. 

mMitra~\cite{mMitra} is one such program conducted by the NGO ARMMAN~\cite{ArmmanFoundation}. mMitra is the second largest maternal mHealth program in the world. The program enrols expectant and new mothers from underserved communities in Mumbai, Palghar, Pune and Nashik in India. Once enrolled, the beneficiary receives pre-recorded automated voice calls every week throughout pregnancy and until one year post child birth. These messages are timed according to their gestational age and contain critical health information around pregnancy, ante-natal and post natal care, child care practices, family planning and many other relevant critical topics. In fact, ARMMAN validated that regular listenership of mMitra 
automated voice call 
messages
has a significant positive impact on not only improved awareness and self care and infant care practices among mothers, but also led to improved health outcomes for both the mothers and babies \cite{hegde2016assessing,murthy2020effects}.

However, despite the known benefits and outreach of the program, ARMMAN noticed a decline in beneficiary engagement both in terms of listenership and drop offs from the program. 
Adherence to public health programs is a well known challenge 
~\cite{killian2024new,arogya,51564}.
A typical solution to improve adherence to such programs is the use of targeted interventions.
In particular, interventions such as calls or visits from community health workers can help keep beneficiaries engaged by providing personalized support and encouragement. 
ARMMAN, deployed a similar intervention whereby health workers would reach out to beneficiaries via live service calls to encourage them to listen to messages regularly, as well as explain to them the importance of listening to the messages.
However, due to limited support staff, it is not feasible for the health workers to reach out to every beneficiary, making it essential to prioritize and identify beneficiaries who would benefit the most from receiving an intervention at a given time.

This is essentially a limited resource allocation problem which involves complex sequential decision making, to identify beneficiaries every week who are at risk of drop off the program and may benefit from receiving an intervention. Restless multi-armed bandits (RMABs)  have been shown to be an effective approach to address such  sequential resource  allocation  problems ~\cite{qian2016restless,liu2010indexability}. In fact, the research team together with ARMMAN developed a RMAB model ~\cite{51909} to identify the beneficiaries that would most benefit from receiving an intervention at a given time. This model was thoroughly evaluated in the lab, followed by a real world field study~\cite{DBLP:conf/atal/VermaM0MHTT23} and then underwent further iterations and finally evolved into an algorithm that was deployed~\cite{verma2023deployed} by the NGO and used on a weekly basis. The model had in fact served $\textbf{350,000}$ mothers until $2024$. 
Despite, the RMAB model having demonstrated clear gains in listenership and reduced drop-off rates in ~\cite{DBLP:conf/aaai/MateMTMVSHVT22,51909}, a crucial gap remained in understanding whether this improved engagement translated into better knowledge comprehension and, consequently, changes in health improving behaviors.



Therefore, the primary objective of this paper is to demonstrate the impact of enhanced engagement due to AI-predicted interventions into measurable knowledge gains and facilitating improvements in behavioral and health outcomes. To achieve this, in collaboration with the NGO, we designed and conducted a randomized controlled trial (RCT), with two arms, and a total of over $34,000$ beneficiaries. The intervention arm received additional support through interventions predicted by the RMAB model, while the control arm did not. Both groups, however, continued to receive standard weekly automated voice calls. The impact of the AI interventions was determined by comparing the knowledge and behavioral outcomes between these two groups of beneficiaries.


Although numerous commendable public health programs exist that enhance adherence and awareness across a variety of health topics ~\cite{arogya,51564,killian2024new}, 
few~\cite{DBLP:journals/corr/abs-2009-09559} have been able to establish the impact of AI interventions on health outcomes.
In this paper, we establish statistically significant improvements in both the awareness and health behaviors of beneficiaries due to the AI-targeted interventions (Table~\ref{table:improvement}), e.g., our AI (RMAB) interventions led to significantly enhanced iron and calcium intake post-delivery. 

An additional major  contribution of this paper is the experimental methodology for end-to-end evaluation of health outcomes of AI interventions. Careful evaluation of AI interventions in the field is extremely time consuming, and in our case, given sequential decision making in the interventions, how to conduct such an evaluation is itself not clearly known. We present a novel approach, in \Cref{sec:sec_setup}, which includes careful comparison of the Whittle index based interventions in restless bandits with corresponding “dummy Whittle index” based control group for reliable and robust counterfactual reasoning. This requires a thorough and detailed experimental methodology which we present in detail in our paper.

Moreover, while this study was conducted within the specific context of the mMitra program, 
we believe the demonstrated success in converting increased engagement into measurable knowledge and behavioral gains offers a valuable model. Similar public health programs, whether focused on preventative care, managing chronic diseases, or promoting lifestyle modifications, could harness AI-targeted interventions to improve  participant engagement and knowledge retention,
and ultimately, the adoption of healthier behaviors and achievement of better health outcomes 
thereby amplifying their impact on public health, particularly across marginalized communities with limited access to resources.



\begin{table*}[t]
\centering
\scalebox{1.5}{
\begin{tabular}{|l|l|l|}
\hline
\rowcolor[HTML]{FFCE93} 
\textbf{Question}                            & \textbf{Percentage improvement} & \textbf{p-Value} \\ \hline
\rowcolor[HTML]{FFFFC7} 
``Did you take iron pills after delivery?''    & \textbf{21.74 \%}               & \textbf{0.0981}  \\ \hline
\rowcolor[HTML]{FFFFC7} 
``Did you take calcium pills after delivery?'' & \textbf{28.00 \%}               & \textbf{0.0413}  \\ \hline
\rowcolor[HTML]{FFFFC7} 
``What was the baby's weight at birth?''       & \textbf{8.97\%}                 & \textbf{0.0080}  \\ \hline
\end{tabular}
}
\caption{Key improvement on health knowledge and behaviour  questions due to AI (RMAB) interventions  with  p-Values}
\label{table:improvement}
\end{table*}

\section{Related Work and Background} \label{sec:sec_prev_work}

\textbf{RMABS}: The allocation of finite resources is a recurring challenge across various domains necessitating strategic planning. Restless multi-armed bandits (RMABs) serve as a prevalent instrument for addressing such sequential resource allocation problems within uncertain environments and have demonstrated their utility in applications such as anti-poaching surveillance \cite{qian2016restless}, multi-channel communication optimization \cite{liu2010indexability}, task scheduling \cite{bagheri2015restless, yu2018deadline}, unmanned aerial vehicle routing \cite{zhao2008myopic}, and others.
In collaboration with ARMMAN, \citet{DBLP:conf/aaai/MateMTMVSHVT22} presented a RMAB based approach 
to determine the allocation of service calls, where each beneficiary is modeled as a Markov decision process. This initial model was evaluated thoroughly 
in a field study~\citet{DBLP:conf/aaai/MateMTMVSHVT22} and deployed~\cite{verma2023deployed}. 

\paragraph{Decision Focused Learning (DFL):} While the above approach optimized for predictive accuracy to learn model parameters, followed by solving a separate optimization problem using the learnt parameters. Decision Focused Learning on the other hand, optimizes for decision quality~\cite{shah2022decision}, by integrating these two stages into an end-to-end system. In particular, focusing on how well predictions lead to effective actions rather than solely on the accuracy of the predictions themselves. The previous model with ARMMAN was then evolved into a  DFL based model~\cite{51909} to improve the listenership in the mMitra program via DFL predicted interventions which was again evaluated in a field study and eventually \textbf{deployed} ~\cite{DBLP:conf/atal/VermaM0MHTT23}. 

\paragraph{Whittle Index} is a heuristic policy tool \cite{whittle1988restless} used primarily for solving RMABs, which are notoriously complex sequential resource allocation tasks. 
The Whittle Index assigns a numerical value (the index) to each arm, representing the benefit of acting on that arm, naturally providing a ranking or prioritization of which arms should be acted on for maximal benefit,
following the budget constraints.

\paragraph{Behavioral and Health Outcomes:} To date, the primary observable objective optimized by the deployed model and similar intervention scheduling programs for ARMMAN has been the mother's listenership of automated voice calls; consequently, program performance has consistently been evaluated through improvements in listenership metrics. However, a correlation between AI-scheduled interventions and behavioral outcomes had not been demonstrated, until \cite{dasgupta2024preliminary} which showed a preliminary investigation and positive trends in some questions in a health study. However, it failed to establish concrete statistical significance due to several limitations. This paper establishes more conclusive results 
by improving upon the methodology of conducting the survey, asking more targeted questions, as well improvising the evaluation methodology via a more robust and reliable counterfactual comparison.


\section{Study Setup} \label{sec:sec_setup}

Our initial methodological step involved segmenting registered beneficiaries into distinct cohorts, categorized by their respective program enrollment dates. Within each cohort, we then implemented a randomized allocation procedure, creating intervention and control groups. It is important to note that all participants, regardless of group assignment, continued to receive the automated voice messages every week containing essential health information 
as per schedule. However, only those assigned to the intervention group were considered for supplementary interventions. The DFL algorithm played a crucial role in determining, on a weekly basis, which intervention group members would receive personalized live service calls from health workers. Finally, to evaluate the impact of these interventions on behavioral and health knowledge, the NGO conducted a comprehensive survey on representative subsets of both the intervention and control groups.


\subsection{Cohorts}
The study was conducted in three cohorts with a combined number of $34453$ beneficiaries, with details on the number of enrolments and respective timelines in ~\cref{table:cohort}. As explained later in \Cref{subsub:int}, the program treats these cohorts as entirely distinct groups. And they also serve primarily as a mechanism for establishing intervention eligibility timelines.

\begin{table*}[t]
\centering
\scalebox{1.5}{

\begin{tabular}{|c|c|c|c|}
    \hline
    Cohort & Beneficiaries Registered & Date of Registration &
    Timeline of Intervention\\
    \hline
     $1$ & $12749$ & $1$st Oct to $31$st Oct $2023$ &
    $5$th Feb to $3$rd Mar $2024$\\
    \hline
     $2$ & $9122$ & $1$st Nov to $30$th Nov $2023$ &
    $4$th Mar to $24$th Mar $2024$\\
    \hline
     $3$ & $12582$ & $1$st Dec to $31$st Dec $2023$ &
    $25$th Mar to $21$st Apr $2024$
    \\
    \hline
\end{tabular}
}
\caption{Registration and Intervention Timelines and details of the three cohorts }
\label{table:cohort}
\end{table*}


\subsection{Experiment Arms}

Within each cohort, beneficiaries were randomly assigned to either the intervention or control arm, ensuring a similar distribution of key attributes between the two groups as described below. 
This procedure mirrors covariate adaptive randomization, a technique designed to balance the distribution of relevant covariates across experimental groups~\cite{lachin1988randomization}. We specifically balanced the distribution across arms based on gestational age to ensure similar distribution of expectant women and women post birth, as well as listenership behaviours as described below.

\begin{description}
    \item[Engagement States] \phantom{a}
    \begin{itemize}
        \item For each beneficiary and a given automated voice call, we denote the engagement state $E@T$ at a threshold $T$ as $E@T = 1$ if the beneficiary listened to the call for at least $T$ seconds, and $E@T = 0$ otherwise.
        \item We compute $E@T\_w$ for each beneficiary, representing the engagement state over $w$ weeks leading up to the cohort's anticipated intervention start date.
        \item To ensure comparable listenership patterns between the two arms, we aim to achieve approximately equivalent values of $E@T\_w$ for thresholds $T\in \{1, 5, 10, 30, 100\}$ and time windows $w\in \{1,2,3\}$.
    \end{itemize}
    
\end{description}

Based on previous studies and discussions with the NGO \cite{verma2023deployed}, we observed it was critical to ensure a balanced distribution of above attributes (based on gestational age and  $E@T\_w$) to avoid any biases in the cohort distribution and ensure a fair evaluations of both arms. This was achieved, by constructing a feature vector $Y$ for each beneficiary by concatenating the above attributes. Subsequently, we partition the beneficiaries into two equally sized groups, employing $Y$ as the stratification criterion \cite{51909}. This is accomplished by treating $Y$ as a categorical label and utilizing a stratified splitting mechanism to produce two balanced subsets. Specifically, we leverage the stratified option within the \textit{train\_test\_split} function from the sklearn library \cite{pedregosa2011scikit}. Given the sufficient number of beneficiaries within each cohort, we successfully achieve an exact split, resulting in perfectly balanced groups.


\subsection{Conducting AI predicted Interventions}
Interventions in mMitra are conducted via live service calls made by healthcare workers to encourage regular listenership of the automated voice messages.

\subsubsection{Number of Interventions per week.}\label{subsub:int}
Interventions began on February $2024$. Interventions were only given to beneficiaries that had been present for at least $3$ weeks in the program. We perform interventions on beneficiaries of cohort $1$, $2$ and $3$ one after the other, intervening on a total of ~$35\%$ of beneficiaries in each cohort to simulate the realistic scenario of intervening on only a subset of the beneficiaries, with exact respective timelines of each cohort detailed in Table~\ref{table:cohort}.


The NGO discouraged repeated calls if a woman’s listening behavior remains unchanged. This consideration makes the model more aligned with real-world human behavior, ultimately leading to a more effective and respectful intervention strategy. Hence, both in the deployed model and this study each beneficiary can be intervened on only once. 
 
 The NGO conducted approximately $330$ interventions per week for cohorts $1$ and $3$ and about $285$ for Cohort $2$. This was done to keep the approximate proportion of interventions similar for each week, and since Cohort $2$ had lesser registrations, the number of weeks is lesser. In total about $~12000$ interventions were conducted.

\subsubsection{Eligibility for Interventions.}
Beneficiaries are deemed eligible for interventions when the following criteria are met:
\begin{enumerate}
    \item Active Program Status: They are currently receiving automated voice messages.
    \item Recent Engagement: They have answered at least one automated voice message within the four weeks preceding the intervention period (to ensure phone number is still active/valid).
    \item No Prior Intervention: They have not previously received an intervention.
\end{enumerate}

Based on our discussion with the NGO, these eligibility requirements serve to ensure the efficient and equitable allocation of the program's constrained intervention resources.

\subsubsection{Conducting Interventions.}

Each week, the DFL \cite{51909} algorithm, determines the set of beneficiaries from the intervention arm who will receive an intervention, 
which we refer to as the list $I_D$.
We also simulate the DFL algorithm on the control (dummy) arm to determine the set of beneficiaries that would have been selected for an intervention (assuming we conducted the same number of interventions as in the intervention arm). As in the intervention arm, beneficiaries in the control arm cannot be selected multiple times. These beneficiaries from the control arm 
are stored into a list $I_C$. The intervention list $I$ refers to the combined lists $I_D$ and $I_C$. The idea behind this is that we later compare the behavior of beneficiaries from $I_D$ and $I_C$, as we can think of beneficiaries from $I_C$ as the counterfactual counterparts of those from $I_D$, inspired by \citet{boehmer2024evaluating} which established this method of evaluating index-based policies in a statistically meaningful way. The interventions concluded on $21$st April $2024$.

\subsection{Health Knowledge and Behavioral Assessment Process}

\subsubsection{Survey Questions Design and Objective}

The health knowledge and behaviour assessing survey contained $23$ questions (\Cref{tab:scores_12_sc} and \Cref{tab:scores_12_mc}), designed by the NGO to evaluate their 
knowledge across various health-related topics. This assessment is aligned with the content of the automated voice messages, aiming to gauge the beneficiaries' comprehension of the delivered information. Specifically, the survey encompassed categories such as program engagement, breastfeeding, micro nutrient supplementation, perinatal care, early infancy care, family involvement in maternal health and perceived benefits of the program. As generic maternal and child health information is widely available, the survey focused on those messages which are unique to the mMitra program. \Cref{tab:scores_12_mc} and  \Cref{tab:scores_12_sc} provide a comprehensive list of these questions. Some questions have an Ante-natal Care (ANC) equivalent which implies asking about care pre-delivery. For each question, beneficiaries received a score based on their responses. 

Among the $23$ questions, $13$ were structured as single-choice (Yes/No) questions. The remaining $8$ were questions where scores were determined based on multiple possible correct answers, with varying weights assigned to different answers. Beneficiaries were offered the choices, and could select one or more of these answers, and their final score was the cumulative score of all correctly identified answers within that question.

Again, a key difference compared to the previous study (\cite{dasgupta2024preliminary}) was the way the questions were structured by the NGO, to reduce ambiguity in questions and responses, thereby ensuring a more reliable way to quantitatively evaluate performance across the two groups.

\subsubsection{Conducting the Survey} \label{sec:survey_details}

The surveys were then conducted by the NGO on the beneficiaries from the intervention list~$I$ between $20$th June to $30th$ October in $2024$. We also excluded from $I_C$ and $I_D$, the beneficiaries who had given birth before the interventions (to be able to conduct post-natal surveys). Additionally, among of all the people who were scheduled to receive interventions in $I_D$, only a fraction 
actually answered the intervention call (say this set is $I_{D'}$). This combined set of $I_C$ and $I_{D'}$ is then considered for the survey. 

This is a key difference from the previous study \cite{dasgupta2024preliminary} where the authors surveyed and included the entire set $I_D$ irrespective of whether the intervention actually happened or not, 
thereby impacting the reliability of the comparison, since $I_D$ then contained both people who had and had not answered the intervention call.

\begin{table*}[t]
\centering
\scalebox{1}{
\begin{tabular}{@{}|l|l|l|l|l|l|l|@{}}
\toprule
                               & {\color[HTML]{222222} Registered}                                                                                & {\color[HTML]{222222} \begin{tabular}[c]{@{}l@{}}After removing\\ inactive\end{tabular}} & {\color[HTML]{222222} \begin{tabular}[c]{@{}l@{}}Intervened On \\ {[}About 35\%{]} \\ (for Intervention)\end{tabular}} & {\color[HTML]{222222} \begin{tabular}[c]{@{}l@{}}Picked up \\ Intervention \\ (for Intervention)\end{tabular}} & {\color[HTML]{222222} \begin{tabular}[c]{@{}l@{}}After imposing delivery-\\ date constraint \\ (delivery after intervention : \\ check \cref{sec:survey_details})\end{tabular}} & {\color[HTML]{222222} Picked up Survey} \\ \midrule
{\color[HTML]{222222} DFL}     & {\color[HTML]{222222} }                                                                                          & {\color[HTML]{222222} }                                                                  & {\color[HTML]{222222} 4495}                                                                                   & {\color[HTML]{222222} 3469}                                                                           & {\color[HTML]{222222} 1496}                                                                                                                & {\color[HTML]{222222} 701}              \\ \cmidrule(r){1-1} \cmidrule(l){4-7} 
{\color[HTML]{222222} Control} & \multirow{-2}{*}{{\color[HTML]{222222} \begin{tabular}[c]{@{}l@{}}12749 + 9122 + \\ 12582 = 34453\end{tabular}}} & \multirow{-2}{*}{{\color[HTML]{222222} 19970}}                                           & {\color[HTML]{222222} 4495}                                                                                   & {\color[HTML]{222222} 4495}                                                                           & {\color[HTML]{222222} 1901}                                                                                                                & {\color[HTML]{222222} 850}              \\ \bottomrule
\end{tabular}
}
\caption{Beneficiary Counts at Different Stages}
\label{tab:beneficiary_counts}
\end{table*}





In this study, the subset of beneficiaries $I_{D'}$ and $I_C$ are then called by a health worker to answer the questions from the survey. 
Further, to ensure fairness, the interviewers were blinded and had no idea whether the interviewee belonged to Control or intervention group. However, again, the survey calls may not be answered by all beneficiaries.
 This makes it difficult to evaluate the outcome of the study as we know the survey results for only a subset of the beneficiaries that are willing and available to answer to the survey questions (in particular, this group of beneficiaries is not chosen uniformly at random). Hence, we have to re-balance the control and intervention group for the final comparison.
The exact numbers at each stage of the study can be seen in \Cref{tab:beneficiary_counts}.

\begin{table*}[t]
\centering
\scalebox{}{}
\begin{tabular}{|l|l|l|l|r|}
\hline
\rowcolor[HTML]{FFCE93} 
\textbf{Question}                                                                                                                                      & \textbf{Correct Answer}   & \textbf{Control Score} & \textbf{Intervention Score} &
\multicolumn{1}{l|}{\cellcolor[HTML]{FFCE93}p-Value} \\ \hline
\rowcolor[HTML]{ECF4FF} 
    Did you know your baby's birth weight ?                                                                                                       & Yes                  & 0.991 $\pm$ 0.004 & 0.993 $\pm$ 0.003      & 1                                                    \\ \hline
\rowcolor[HTML]{ECF4FF} 
How many iron tablets did you consume during pregnancy?                                                                                                                          & Recommended or more  & 0.836 $\pm$ 0.017 & 0.827 $\pm$ 0.018      & 0.776818                                             \\ \hline
\rowcolor[HTML]{ECF4FF} 
How many calcium tablets did you consume during pregnancy?                                                                                               & Recommended or more  & 0.836 $\pm$ 0.017 & 0.834 $\pm$ 0.018      & 0.99313                                              \\ \hline
\rowcolor[HTML]{ECF4FF} 
What is the name of the pill to be taken for preventing anemia?                                                                                                                     & Iron Pill                  & 0.457 $\pm$ 0.023 & 0.431 $\pm$ 0.023      & 0.473308                                             \\ \hline
\rowcolor[HTML]{ECF4FF} 
What is the name of the pill to be taken for strengthening bones?                                                                                                                  & Calcium Pill                  & 0.494 $\pm$ 0.023 & 0.497 $\pm$ 0.024      & 0.984697                                             \\ \hline
\rowcolor[HTML]{ECF4FF} 
\textbf{Are you currently taking iron tablets after delivery?}                                                                                         & \textbf{Yes}                  & \textbf{0.234 $\pm$ 0.019} & \textbf{0.283 $\pm$ 0.021}      & \textbf{0.098111}                                             \\ \hline
\rowcolor[HTML]{ECF4FF} 
\textbf{Are you still taking calcium pills after delivery?}                                                                                            & \textbf{Yes}                  & \textbf{0.244 $\pm$ 0.019} & \textbf{0.306 $\pm$ 0.022}      & \textbf{0.041262}                                             \\ \hline
\rowcolor[HTML]{ECF4FF} 
When you faced anxiety, did you share it with your husband?                                                                                                        & Yes                  & 0.1 $\pm$ 0.014   & 0.106 $\pm$ 0.015      & 0.877882                                             \\ \hline
\rowcolor[HTML]{ECF4FF} 
\begin{tabular}[c]{@{}l@{}}Should the baby be fed the first solid yellow \\ milk that comes to the mother after childbirth?\end{tabular}      & Yes                  & 0.881 $\pm$ 0.015 & 0.87 $\pm$ 0.016       & 0.665967                                             \\ \hline
\rowcolor[HTML]{ECF4FF} 
Should the baby be fed at night?                                                                                                              & Yes                  & 0.982 $\pm$ 0.006 & 0.962 $\pm$ 0.009      & 0.102588                                             \\ \hline
\rowcolor[HTML]{ECF4FF} 
How does the baby react when you talk to it (at 3 months)                                                                                                                                  & Gives a Social Smile & 0.697 $\pm$ 0.021 & 0.724 $\pm$ 0.021      & 0.406006                                             \\ \hline
\rowcolor[HTML]{ECF4FF} 
Have you heard calls from mMitra regularly?                                                                                                   & Yes                  & 0.713 $\pm$ 0.02  & 0.728 $\pm$ 0.021      & 0.662864                                             \\ \hline
\rowcolor[HTML]{ECF4FF} 
\begin{tabular}[c]{@{}l@{}}Have you ever discussed with your husband/family\\  about the information you heard/told in the call?\end{tabular} & Yes                  & 0.258 $\pm$ 0.02  & 0.274 $\pm$ 0.021      & 0.633354                                             \\ \hline
\end{tabular}
\caption{Scores on single correct questions along with expected answers, standard error and p-values for cohorts 1 and 2.}
\label{tab:scores_12_sc}
\end{table*}

\begin{table*}[]
\centering
\begin{tabular}{|l|l|l|l|r|}
\hline
\rowcolor[HTML]{FFCE93} 
\textbf{Question}                                                                                                                                            & \textbf{Correct Answer (s)}                                                                                                                                                                                                                                     & \textbf{Control Score} & \textbf{Intervention Score} & \multicolumn{1}{l|}{\cellcolor[HTML]{FFCE93}\textbf{p-Value}} \\ \hline
\rowcolor[HTML]{ECF4FF} 
\begin{tabular}[c]{@{}l@{}}\textbf{If yes what was the baby's weight at birth?} \\ \textbf{(Follow up to knowing weight)}\end{tabular}                                         & \textbf{'It was 2.5' or 'It was more than 2.5'}                                                                                                                                                                                                                          & \textbf{0.766} $\pm$ \textbf{0.019}          & \textbf{0.836} $\pm$ \textbf{0.018}               & {\color[HTML]{000000} \textbf{0.007991}}                               \\ \hline
\rowcolor[HTML]{ECF4FF} 
\begin{tabular}[c]{@{}l@{}}If yes, what did the husband do? \\ (Follow up to telling husband about anxiety)\end{tabular}                                     & 'The husband took her to the doctor'                                                                                                                                                                                                                            & 0.059 $\pm$ 0.011          & 0.074 $\pm$ 0.012               & {\color[HTML]{000000} 0.367425}                               \\ \hline
\rowcolor[HTML]{ECF4FF} 
When should the baby be fed the first milk after birth?                                                                                                      & 'Within 1 hour'                                                                                                                                                                                                                                                 & 0.633 $\pm$ 0.022          & 0.645 $\pm$ 0.023               & {\color[HTML]{000000} 0.709437}                               \\ \hline
\rowcolor[HTML]{ECF4FF} 
What to do to dry baby's umbilical cord after birth?                                                                                                         & \begin{tabular}[c]{@{}l@{}}'It should be kept clean and dry' or\\ 'Nothing should be applied'\end{tabular}                                                                                                                                                      & 0.545 $\pm$ 0.023          & 0.575 $\pm$ 0.023               & {\color[HTML]{000000} 0.353922}                               \\ \hline
\rowcolor[HTML]{ECF4FF} 
\begin{tabular}[c]{@{}l@{}}If yes, what made you want to listen to the call? \\ (Follow up to question about if they listen \\ to mMitra calls)\end{tabular} & \begin{tabular}[c]{@{}l@{}}'Doubts clarified',  'Endorsement', \\ 'Guidance',\\ 'Knowledge of physiological changes'\\ 'Fills gaps left by medical advice'\end{tabular}                                                                                         & 0.32 $\pm$ 0.013           & 0.331 $\pm$ 0.014               & {\color[HTML]{000000} 0.596208}                               \\ \hline
\rowcolor[HTML]{ECF4FF} 
\begin{tabular}[c]{@{}l@{}}Does anyone in your family know that you are getting \\ informative calls from mMitra?\end{tabular}                               & \begin{tabular}[c]{@{}l@{}}'Husband', 'In-laws', \\ 'Mother', 'Relative'\end{tabular}                                                                                                                                                                           & 0.844 $\pm$ 0.016          & 0.856 $\pm$ 0.017               & {\color[HTML]{000000} 0.611126}                               \\ \hline
\rowcolor[HTML]{ECF4FF} 
Do they hear these calls too?                                                                                                                                & \begin{tabular}[c]{@{}l@{}}'Yes , always',\\ 'Yes, I put the mobile on speaker \\ so that everyone in the family \\ can hear the call', \\ 'yes , sometimes' (partial points)\\ 'No , but I tell them what \\ the call was about' (partial points)\end{tabular} & 0.302 $\pm$ 0.014          & 0.293 $\pm$ 0.015               & {\color[HTML]{000000} 0.663034}                               \\ \hline
\rowcolor[HTML]{ECF4FF} 
\begin{tabular}[c]{@{}l@{}}Do you pick up mMitra calls even when other people \\ are around?\end{tabular}                                                    & \begin{tabular}[c]{@{}l@{}}'Yes', 'Always', \\ 'Sometimes' (partial points)\end{tabular}                                                                                                                                                                        & 0.538 $\pm$ 0.02           & 0.524 $\pm$ 0.021               & {\color[HTML]{000000} 0.630469}                               \\ \hline
\rowcolor[HTML]{ECF4FF} 
\begin{tabular}[c]{@{}l@{}}Do you pick up mMitra call even when you are \\ busy at work?\end{tabular}                                                        & \begin{tabular}[c]{@{}l@{}}'Yes', 'Always', \\ 'Sometimes' (partial points)\end{tabular}                                                                                                                                                                        & 0.46 $\pm$ 0.019           & 0.467 $\pm$ 0.021               & {\color[HTML]{000000} 0.795084}                               \\ \hline
\end{tabular}
\caption{Scores on single correct questions along with expected answers, standard error and p-values for cohorts 1 and 2.}
\label{tab:scores_12_mc}
\end{table*}

\section{Results} \label{sec:sec_results}

\subsection{Comparison Methodology}
As explained above in \Cref{sec:survey_details}, a direct comparison of the people finally surveyed (control vs intervention) does not suffice for comparison, since not everyone chosen for intervention ($I_D$) and/or survey ($I$), ends up answering the call, and this set of people 
is non-random, as described in \Cref{sec:survey_details}. This motivates 
the need for a more robust and fair comparison between the two groups.


Prior work \cite{dasgupta2024preliminary} selected survey respondents from the intervention group ($I_D$) and formed a matched control group by minimizing Mahalanobis distance using manually selected features like pre-intervention listenership, gestational age, and number of previous children. Firstly, this approach doesn't account for imbalances from those who never answered the intervention call potentially diluting the impact of the interventions (as highlighted in ~\ref{sec:survey_details} and secondly, they employed handcrafted features for counterfactual matching, which could again potentially impact the comparison's reliability.
Consequently, they were unable to establish conclusive impact of AI interventions on behavioral outcomes.

In this study, however, we instead perform counterfactual matching of beneficiaries using the Whittle Index \cite{whittle1988restless}, which is the index or ranking used to determine who gets an intervention 
by the DFL-RMAB problem, guiding resource allocation by prioritizing arms that offer the most expected marginal reward. 
The matching is done by greedily matching the Whittle index across the control and intervention arms, making sure that the difference is below a threshold of $0.01$.

Whittle indices are computed both for intervention and control group, but the actual interventions only go out to the the intervention group. For the control group, the whittle indices only help us identify beneficiaries who would have been intervened on by the model had we conducted interventions on that group. The advantage of using this mechanism is that, firstly, \citet{boehmer2024evaluating} demonstrated, this approach to be a statistically valid mechanism of evaluating a RCT involving an index based policy.
Furthermore, the Whittle index generated by the DFL algorithm~\cite{51909} inherently learns and weighs relevant demographic features and listenership behaviors in the prediction process. Given that interventions are ultimately assigned by ranking these Whittle indices, the index value itself signifies the algorithm's preference for intervening on a particular beneficiary. This implies a consistent preference for beneficiaries with similar index values, regardless of whether they are in the intervention or control group. Therefore, to ensure a robust and reliable counterfactual comparison, beneficiaries in the intervention set ($I_D'$) who answered the survey were matched with a beneficiary with a similar Whittle index in the control set ($I_C$) in the same cohort and who answered the survey. This is a \textbf{key methodological improvement that led to a more stable and reliable counterfactual comparison}.



\subsection{Key Results}

\subsubsection{Establishing Improved Listenership and Engagement}

Towards the goal of assessing improvement in knowledge and behavioral outcomes across the two groups, we first evaluate whether the intervention successfully enhanced listenership across the two groups. The gain in listenership is quantified as the difference between post-intervention listenership and pre-intervention listenership, 
which is calculated by taking the average listenership per user over eight weeks post intervention and per user over eight weeks
pre-intervention 
respectively. It was established by the NGO and prior work \citet{dasgupta2024preliminary, verma2023deployed}, that listenership typically gradually reduces over time, motivating the need for interventions. 

\begin{figure*}[htbp]
    \centering
    \includegraphics[width=0.6\textwidth]{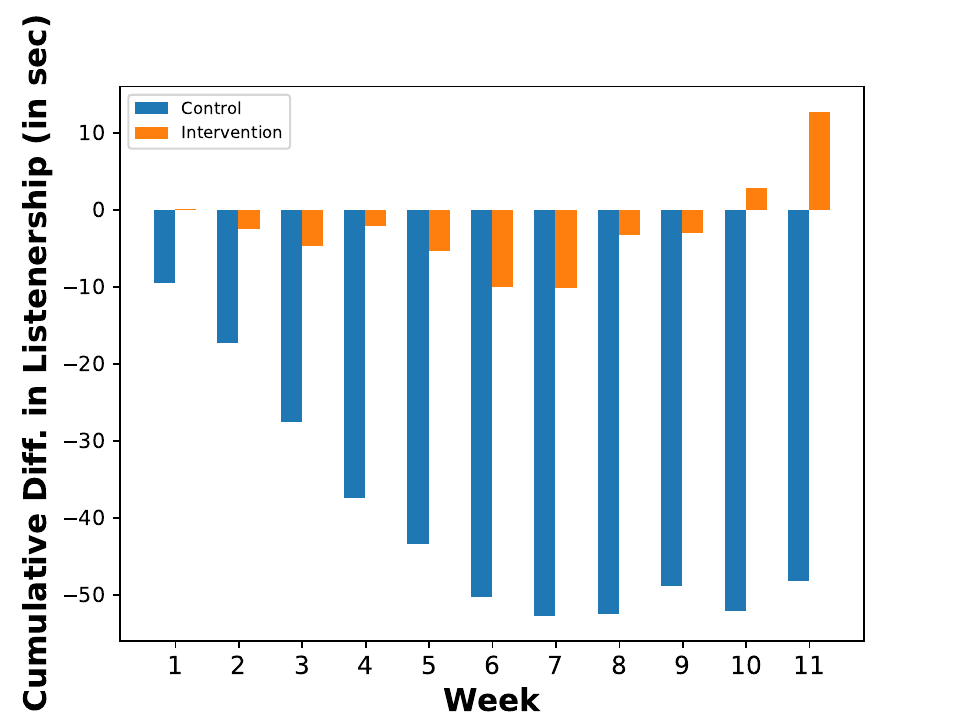}
    \caption{Comparison of \textbf{Weekwise Listenership Gain} by comparing listenership behaviours post vs pre intervention for both Intervention and Control for all Cohorts. Drop in Intervention Group is significantly lower than Control indicating the impact of interventions in improving listenership behaviours of the Intervention Group.}
    \label{fig:listenership_gain}
\end{figure*}

We observe the same in Figure~\ref{fig:listenership_gain} which shows a general trend of decreasing listenership in the control group by $8.5\%$, in comparison to the intervention arm which gains by $3.6\%$, establishing that interventions led to reduced drop in listenership.
The x-axis represents the week number, indicating the point in time at which the listenership gain is calculated. The y-axis represents the cumulative listenership gain in seconds, computed as the cumulative sum of the mean listenership gain for all beneficiaries who were intervened (or would have been intervened) in a given week. This approach ensures that for each week, we only consider the listenership gain of the relevant beneficiaries, providing a clear week-wise trend of the intervention's impact.

While listenership gradually reduces over time across both groups (as also shown in \cite{verma2023deployed, DBLP:conf/aaai/MateMTMVSHVT22}), however, as seen in Figure~\ref{fig:listenership_gain}, across all weeks, we observe a general trend where the intervened group's listenership drop is significantly lesser than that of the control group. 

Interestingly, Cohort 3 shows an overall improvement in listenership in the control group
compared to Control group of Cohorts 1 and 2. However, the gain still remains lower than the corresponding Cohort 3 in Intervention group, showing that the interventions have still had a significant positive impact on listenership among all cohorts. 

It must be noted that the final goal of this study is to establish the benefit of the AI predicted interventions on health and behavioral outcomes. Hence, the comparison here is done only against the baseline with no-intervention. The efficacy of different AI methodologies in predicting interventions for improving listenership was already demonstrated in \cite{DBLP:conf/aaai/MateMTMVSHVT22} and \cite{51909} and DFL was shown to significantly outperform several baselines and was hence deployed and used in this study.

\subsubsection{Establishing Impact on Health Knowledge and Behavioral Outcomes}

The final objective of the study is to assess the improvements in health related knowledge and behaviours due to the AI predicted interventions. Hence, the responses to the survey questions were compared across the single and multiple choice questions across both the intervention and control group. 
Key findings from the study are listed below, focusing in particular, on the improvement in health-related knowledge and behaviours among beneficiaries in Cohorts 1 and 2.

\begin{figure*}[ht!]
    \centering
    \begin{subfigure}{0.4\textwidth}
        \centering
        \includegraphics[width=\textwidth]{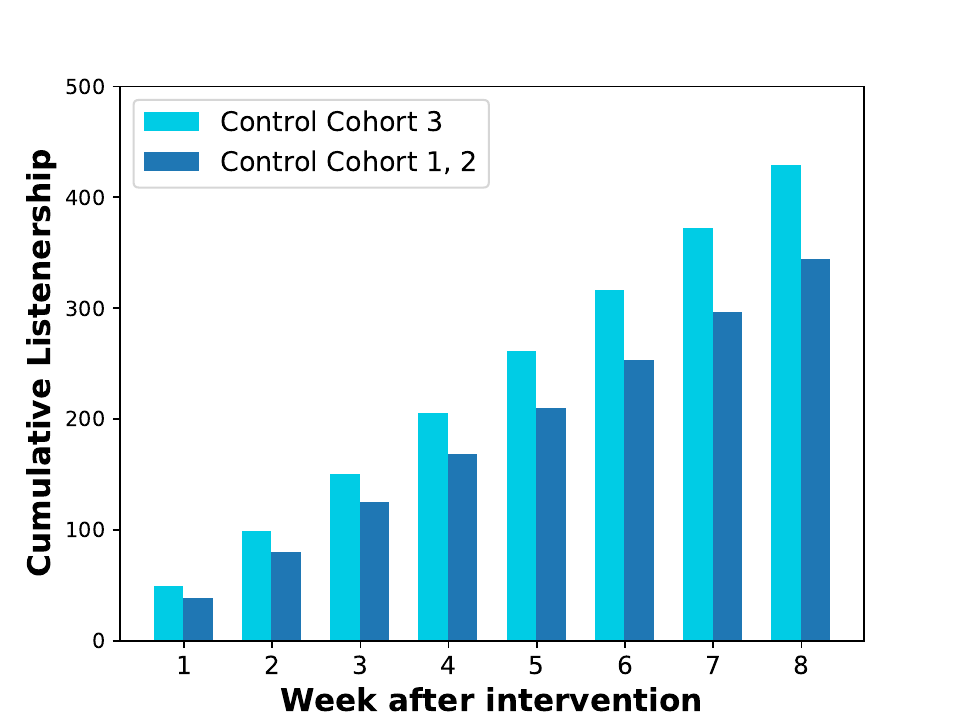}
        \label{fig:cohort_list_dummy}
    \end{subfigure}
    \hfill
    \begin{subfigure}{0.4\textwidth}
        \centering
        \includegraphics[width=\textwidth]{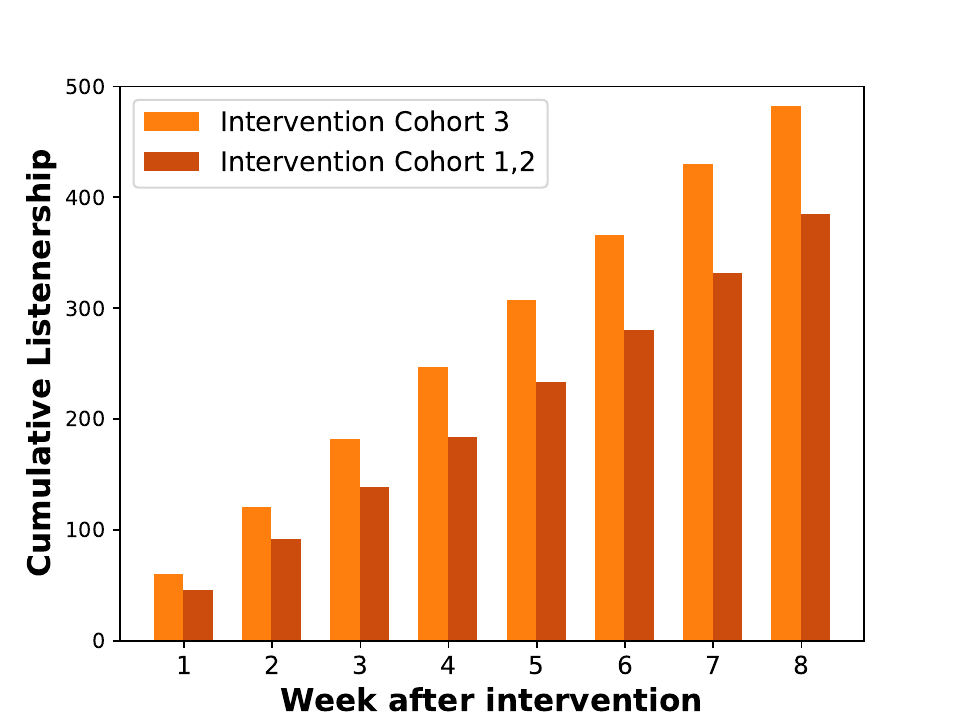}
        \label{fig:cohort_list_dfl}
    \end{subfigure}
    \caption{Comparison of the cumulative average listenership of \textbf{Cohort 1 and 2 vs Cohort 3} after "intervention" for Control group (left) and Intervention Group (right). Both the plots indicate that the \textbf{base listenership for Cohort 3 was higher than Cohort 1 and 2}.}
    \label{fig:listen}
\end{figure*}

The complete list of comparison of response scores from all single choice and multiple choice questions across the intervention and control group for Cohort 1 and 2 can be found in Table ~\ref{tab:scores_12_sc} and Table~\ref{tab:scores_12_mc} respectively, and for all cohorts combined in
the Appendix. While we observe positive trends in several questions, many of these outcomes are not conclusive due to statistical insignificance. However, below we highlight the results for three questions, for which we observed statistically significant improvements in the intervention group.

\paragraph{Key Findings}
\begin{enumerate}
    \item \textbf{"Are you still taking iron pills after delivery?" }
    \item \textbf{"Are you still taking calcium pills after delivery?"} 
    \item \textbf{"What was the baby's weight at birth?"} 
\end{enumerate}

\paragraph{\textbf{Quantitative analysis}}

The improvements on these questions are shown in 
in the Appendix, where the x-axis represents the week, while the y-axis denotes scores on that question. Here, cumulative weeks refer to the inclusion of all beneficiaries who received interventions up to a particular week. For example, in Cumulative Week 3, we consider all beneficiaries who were intervened in Weeks 1, 2, and 3. Similarly, for the control (dummy) arm, although no actual intervention takes place, we calculate scores on the beneficiaries identified in $I_C$ until that week, plotted against the equivalent intervened arm beneficiaries from $I_{D'}$. The Y-axis represents the average score for a particular question corresponding to the respective groups. This score is derived from beneficiary responses to the specific knowledge-based questions asked. 

Furthermore, as seen in figures attached in 
the Appendix, we observe that towards the end of Cohort 2, the cumulative score differences between the intervened and control groups increase, with the intervened group performing better. These results become more evident as we also observe a significant decrease in the p-values for Cumulative Weeks 6 and 7. Table~\ref{table:improvement} further shows the overall improvement on these three questions and the corresponding p-value achieved by week 7, emphasizing the statistical significance on these findings.

\paragraph{\textbf{Significance of these findings}}
\begin{itemize}
    \item \textbf{Calcium and Iron supplementation:}
    Improved performance on the questions regarding the continued intake of iron and calcium supplements after delivery—is particularly significant as it indicates enhanced knowledge and adherence to crucial postnatal supplementation guidelines. It has been noted that adherence to micronutrient supplementation in the postnatal period is very low compared to the antenatal period, primarily due to lack of counselling by health workers.  Hence, this improved understanding is paramount for safeguarding maternal health during the demanding postpartum period \cite{brown2021physiological}. For instance, consistent calcium supplementation is vital as it plays a key role in replenishing maternal bone mineral density, which can be depleted during pregnancy and lactation
    \cite{chan2004calcium}. Similarly, adequate iron intake is critical; iron deficiency after childbirth not only negatively impacts the mother's cognitive functions, potentially affecting her ability to care for herself and her infant, but can also have detrimental effects on the
    baby's overall growth
    \cite{beard2005iron}. 
    \item \textbf{Knowing the birth weight of the baby}: is significant because this information indicates if the mother has monitored a key health metric from birth. Birth Weight is a precise marker for the infant's health at birth and special home-care measures are required for infants who have been born with low birth weight. Hence, mother's ability to report birth weight is a critical indicator of knowledge \cite{smitha2024compliance}. As \citet{who2006neonatal} emphasizes, birth weight is a critical indicator for neonatal survival and for broader public health monitoring. Moreover, when a mother knows the precise birth weight, it facilitates more effective health worker follow-ups, assisting them in monitoring the child's subsequent health and development. 
\end{itemize}
Therefore, higher scores (Table~\ref{table:improvement}) reflecting correct understanding and practice of iron and calcium supplementation and in recalling birth weight demonstrate a meaningful improvement in the mothers' grasp of vital health knowledge and health-promoting behaviors.




\subsection{Impact of Listenership gain on Survey responses for Cohort 3 }

The above findings were established for Cohort 1 and 2. For Cohort 3 however, 
we cannot establish a statistically significant difference between the intervention and control groups. This is primarily due to the reason that Cohort 3 already has a higher listenership on average (~\cref{fig:listen}) than Cohort 1 and 2, even in the Control Group, suggesting that any marginal further increase in listenership may not significantly impact the knowledge gained by the beneficiaries. This also happens because critical health information is repeated across messages at different times, thereby improving the likelihood of awareness around key topics when overall listenership is already good. A combined evaluation across all cohorts is provided in 
the Appendix.

\section{Conclusion}


This paper compellingly demonstrates that AI-based interventions, by successfully enhancing engagement with mHealth programs, can lead to statistically significant improvements in crucial health behaviors among beneficiaries, such as adherence to iron or calcium supplementation and improved understanding of critical health topics relevant to pregnancy and infancy. This not only highlights the potential of AI to drive meaningful advancements in maternal and child health but also illuminates a pathway for other public health programs. The evidence presented, which translates previously established listenership differences into tangible gains in knowledge, suggests that similar AI-driven strategies could be effectively adapted to boost outcomes in diverse health initiatives. For instance, programs targeting chronic disease management, vaccination uptake, Diabetes, nutritional education could leverage tailored AI interventions to improve participant engagement, knowledge retention, and ultimately, adoption of healthier behaviors and achievement of better health outcomes across various populations.

\section{Ethical Considerations}
It is important to highlight that the study was designed together with the NGO, relying on their domain expertise and was approved by the ARMMAN Ethics Review Board. It was registered on the Clinical Trials Registery India (CTRI) - CTRI/2023/02/049472. Verbal consent was obtained from the participants.
The research team at Google only had restricted access to anonymized data and no PII data was used for the study.

\newpage

\bibliography{sample}

\end{document}